\documentclass[journal]{IEEEtran}
\usepackage{graphicx}
\usepackage{amsmath}
\usepackage{mathtools}
\usepackage[linesnumbered,ruled,vlined]{algorithm2e}
\begin{document}
%
\title{Multi-Conditioned Denoising Diffusion Probabilistic Model (mDDPM) for Medical Image Synthesis}
%
%
%

\author{Arjun~Krishna,
        Ge~Wang,~\IEEEmembership{Fellow,~IEEE,}
        and~Klaus~Mueller,~\IEEEmembership{Fellow,~IEEE}

\thanks{Arjun Krishna is in the Computer Science Department, Stony Brook University, Stony Brook, NY 11794 USA (e-mail: arjkrishna@cs.stonybrook.edu).}
\thanks{Ge Wang is with the Biomedical Imaging Center, Center of Biotechnologies and Interdisciplinary Studies, Rensselaer Polytechnic Institute, Troy, NY 12180 USA (e-mail: wangg6@rpi.edu).}
\thanks{Klaus Mueller is with the Computer Science Department, Stony Brook University, Stony Brook, NY 11794 USA (e-mail: mueller@cs.stonybrook.edu).}
}

\maketitle

\begin{abstract}

Medical imaging applications are highly specialized in terms of human anatomy, pathology, and imaging domains. Therefore, annotated training datasets for training deep learning applications in medical imaging not only need to be highly accurate but also diverse and large enough to encompass almost all plausible examples with respect to those specifications. We argue that achieving this goal can be facilitated through a controlled generation framework for synthetic images with annotations, requiring multiple conditional specifications as input to provide control. We employ a Denoising Diffusion Probabilistic Model (DDPM) to train a large-scale generative model in the lung CT domain and expand upon a classifier-free sampling strategy to showcase one such generation framework. We show that our approach can produce annotated lung CT images that can faithfully represent anatomy, convincingly fooling experts into perceiving them as real. Our experiments demonstrate that controlled generative frameworks of this nature can surpass nearly every state-of-the-art image generative model in achieving anatomical consistency in generated medical images when trained on comparable large medical datasets.

\end{abstract}

\begin{IEEEkeywords}
DDPM, Computed Tomography, Generative AI
\vspace{-15pt}
\end{IEEEkeywords}

\section{Introduction}

\IEEEPARstart{G}{reat} strides have been made in deep learning-based medical applications; however, their potential remains constrained by the scarcity of specialized, highly accurate, high-resolution annotated images suitable to robustly train these learning models. To address this limitation, researchers have explored image synthesis to augment the existing datasets, demonstrating that such methods can generate convincingly realistic medical images \cite{Park2021RealisticHB, han2023, krishna2023, krishna2023MP}.

Yet, the generation of phantom images at full resolution with flawless anatomy remains to be a formidable challenge, particularly when incorporating annotations \cite{han2023, krishna2023, krishna2023MP}. This process is prone to introducing anatomical errors as the generation is constrained by these annotations. Existing methods that achieve partial success in generating full-resolution CT images, capable of deceiving radiologists, predominantly rely on applying unconditional state-of-the-art image generative models \cite{Park2021RealisticHB} to large medical datasets. However, these approaches lack purposeful and diversified generative capabilities, merely producing more non-annotated raw medical images adding to the already abundant general datasets.

In this paper, we introduce a methodology that enables the dynamic application of a series of annotations and constraints during the generation process. Our approach simultaneously generates state-of-the-art, full-resolution CT images, passing our Visual Turing Test and exhibiting superior performance compared to other unconditional state-of-the-art image generative models. To our knowledge, our work represents the first endeavor capable of producing full-resolution CT images with accompanying annotations that maintain anatomical accuracy across all clinically relevant Hounsfield Unit (HU) windows.



In our prior work \cite{krishna2023}, we detailed a method for generating unique and diverse annotated CT lung images to construct balanced datasets. It depended on the independent modeling of annotations, with the generative GAN-based models intricately connected to these annotations. 
Recently, DDPMs \cite{ho2020denoising} have emerged as an alternative to traditional image generative models. Given that trained DDPMs sample images through denoising, researchers have devised unique methods to iteratively guide the sampling process towards specific areas of underlying image distributions \cite{chung2023, Choi2021ILVRCM, ho2022}. This approach resembles the conditional generation in GANs but offers the added benefit that such generative models are not tethered to an underlying modality for condition or guidance.

In this paper, we explore a form of conditional generation \cite{Choi2021ILVRCM} and extend it to encompass multiple annotations/conditions simultaneously as guidance and control. We demonstrate that not only does combining such annotations not depreciate synthesis quality, it also surpasses certain state-of-the-art unconditional image generative models. We show that this new method eradicates most anatomical inaccuracies and successfully passes our previously designed \cite{krishna2023} Visual Turing Test.

\section{Methods}

We start out with a large dataset comprising low dose CT images from various scanners and train a DDPM based on the refinements proposed by Nichol et al. \cite{nichol2021improved}. Subsequently, we investigate the sampling strategy of these trained DDPMs suggested by Choi et al. \cite{Choi2021ILVRCM}, and extend it to incorporate multiple conditional or guidance images. Our findings highlight the significance of this strategy for the purpose of synthesizing medical imaging datasets that are not only highly accurate but also annotated. Moreover, as these guidance techniques are not bound by annotations, they can be effectively employed to enhance annotated images featuring rare anatomies and pathology, thereby fostering the development of a more comprehensive and diversified dataset.

\begin{figure}
\centering
\includegraphics[height=4.8cm]{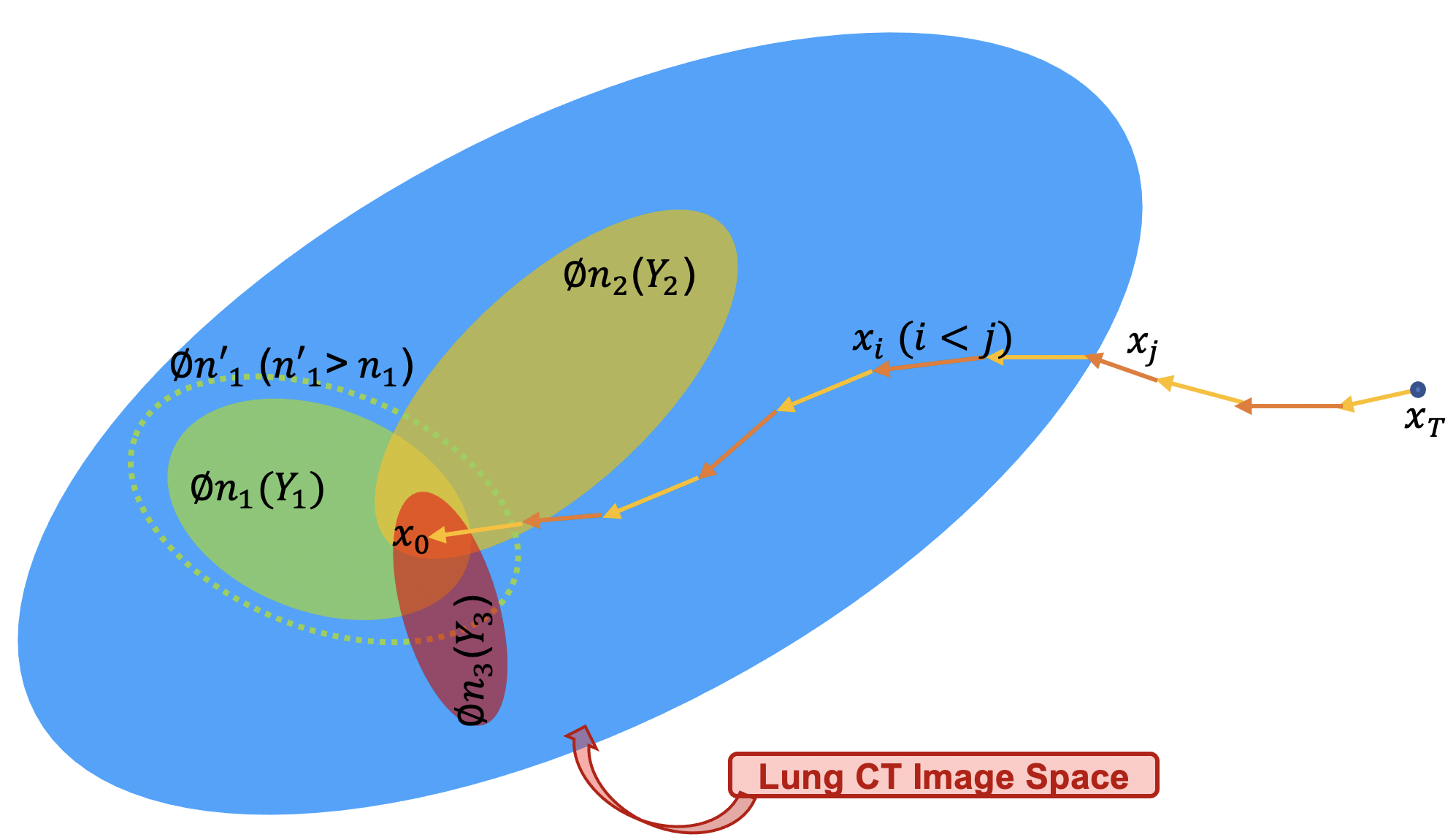}
\caption{Multi-Conditioned Guided Sampling. The blue area represents the image space for all CT lung images; the yellow, green and red circles represent the image space closer to the three guidance images y1, y2 and y3, the size of the circles depends on the images themselves and the downsampling factors n1, n2, n3 of the filter used corresponding to these images.   }
\label{fig}
\vspace{-10pt}
\end{figure}

\subsection{DDPM}

The DDPM we implemented is a Markov Chain model which iteratively converts an isotropic Gaussian distribution into a full Hounsfield window lung CT image data distribution. The Markov Chain model learns the reverse of the forward diffusion process, a fixed Markov Chain that gradually adds noise to the data in the opposite direction of sampling until the signal is destroyed. This forward process is described as:

\vspace{-10pt}

\begin{align}
   q(x_{t}|x_{t-1}) \coloneqq N(x_{t}; \sqrt{1 - \beta_{t}}x_{t-1},\beta_{t}I) 
\end{align}

\noindent where $x_{1}$,...,$x_{T}$ are the latents produced by the addition of noise and $\beta_{1}$,...,$\beta_{T}$ follow a fixed variance schedule. Eq. 1 can be decomposed by the reparameterization trick and $x_{t}$ can be further derived in terms of the image $x_{0}$ as:

\vspace{-10pt}

\begin{align}
   x_{t} = \sqrt{\overline{\alpha}_{t}}x_{0} + \sqrt{1 - \overline{\alpha}_{t}}\epsilon
\end{align}

\noindent where $\alpha_{t}$ $\coloneqq$ 1 - $\beta_{t}$ and $\overline{\alpha}_{t}$ $\coloneqq$ $\prod_{i=1}^t \alpha_i$ and the added noise $\epsilon$ $\sim$ $N(0,I)$ has the same dimensionality as the image and the sampled latents during training. The reverse diffusion process that our model needs to learn is expressed \cite{ho2020denoising} as:

\vspace{-10pt}

\begin{align}
   p_{\theta}(x_{t-1}|x_{t}) = N(x_{t-1}; \mu_{\theta}(x_{t}, t), \sigma^2_{t}I) 
\end{align}

\noindent where $p_{\theta}$ is a neural network to predict $\mu_{\theta}$ and $\mu_{\theta}$ is further decomposed \cite{ho2020denoising} in terms of noise approximator $\epsilon_{\theta}$:

\vspace{-10pt}

\begin{align}
   \mu_{\theta} = \frac{1}{\sqrt{\alpha_{t}}}(x_{t} - \frac{\beta_{t}}{\sqrt{1 - \overline{\alpha}_{t}}}\epsilon_{\theta}(x_{t}, t))   
\end{align}

By formulating the loss function \cite{ho2020denoising} as the log likelihood of $x_{0}$ and computing a variational lower bound (similar to the case of variational auto-encoders) as KL divergence between $q$ and $p$, the authors \cite{ho2020denoising} decided to frame the loss function as the L2 distance between actual mean of the image($\mu$) and $\mu_{\theta}$ which can be further simplified to as the L2 distance between the predicted noise $\epsilon_{\theta}$ and added noise $\epsilon$ at any given time t  

\vspace{-10pt}

\begin{align}
   Loss = \lVert \epsilon - \epsilon_{\theta}(x_{t}, t) \rVert^2 
\end{align}
or
\begin{align}
   Loss = \lVert \epsilon - \epsilon_{\theta}(\sqrt{\overline{\alpha}_{t}}x_{0} + \sqrt{1 - \overline{\alpha}_{t}}\epsilon, t) \rVert^2
\end{align}


Eqs. 2 and 6 are used to train our DDPM, incorporating refinements from Nichol et al. \cite{nichol2021improved}. Our DDPM was trained on a large dataset of 5,000 lung CT scans, with images extracted at full HU width of 2000. This ensures that the generated images span the entire width during sampling and can be visualized at other clinically relevant windows, including lung, bone, and soft-tissue. Utilizing Eq. 3 and the reparameterization trick, $x_{t-1}$ can be sampled as:

\vspace{-10pt}

\begin{align}
   x_{t-1} = \frac{1}{\sqrt{\alpha_{t}}}(x_{t} - \frac{\beta_{t}}{\sqrt{1 - \overline{\alpha}_{t}}}\epsilon_{\theta}(x_{t}, t)) + \sqrt{\beta_{t}}\epsilon  
\end{align}

Using the above equation repeatedly, we can sample lung CT images starting from random noise after training a DDPM on our large dataset. Both training and sampling steps are outlined in prior works related to DDPMs \cite{ho2020denoising}. 
Next, we will focus on the sampling algorithm of our DDPM  to facilitate multi-annotations guidance during our lung CT image generation.

\subsection{Multi-Condition Guidance}

As mentioned earlier, various methods exist for guiding the sampling process of a trained DDPM. In this section, we delve into the guidance techniques presented by Choi et al. \cite{Choi2021ILVRCM} and leverage them for precise control over the generation of lung CT images across all HU windows. Choi et al. posit that it should be feasible to guide the sampling process to a subset of image distributions around a reference image $y$ if we can ensure similarity between the downsampled reference image $y$ and the downsampled generated image $x_{0}$.


\begin{figure*}
\centering
\includegraphics[height=4.9cm]{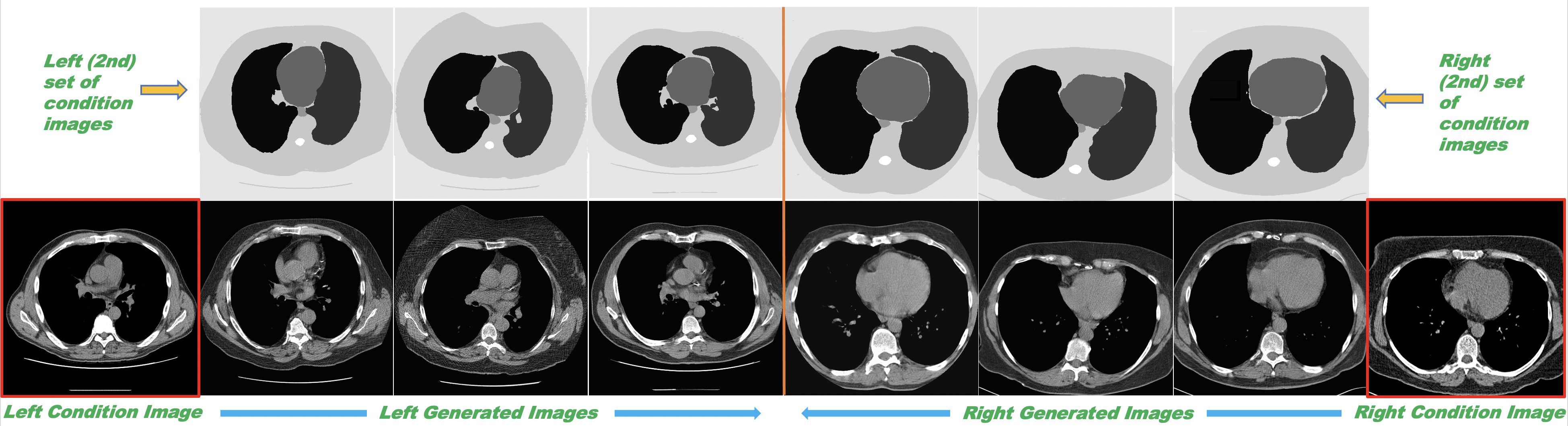}
\caption{This figure shows six examples of lung CT soft-tissue window 2D image generations with two conditional images. Both left and right sections display three generated images for three different anatomy / segmentation maps for the same reference (conditional) CT image, shown in the red boxes. The generations follow the anatomy of the segmentation maps above but exhibit the slice of the heart generation corresponding to the referenced CT images. The results are displayed in the soft-tissue window to highlight the similarity and accuracy of the generated anatomy w.r.t guidance images.}
\label{fig}
\vspace{-10 pt}
\end{figure*}

\begin{algorithm}
  \caption{Sampling}
  \textbf{Input:} Conditional / guidance images $y_{1},....y_{M}$\\
  \textbf{Output:} Generated image x\\
  \textbf{Filter-scales:} $\phi_{n_{1}},....\phi_{n_{M}}$\\
  \textbf{Time-steps (T, a):} $a_{1},....a_{M}$\\
  $x_{T} \sim N(0,I)$ \\
  
  \For{$t=T$ \KwTo $1$}{
    $z \sim N(0,I)$ \\
    \If{t = 1}{
        z = 0\\
      }
    $x_{t-1} = \frac{1}{\sqrt{\alpha_{t}}}(x_{t} - \frac{1 - \alpha_{t}}{\sqrt{1 - \overline{\alpha}_{t}}}\epsilon_{\theta}(x_{t}, t)) + \sigma_{t}z$ \\
    $X = 0$ \\
    \For{$s=1$ \KwTo $M$}{
        $y_{s_{t-1}} \sim q(y_{s_{t-1}}|y_{s})$ \\
        \If{$t \geq a_{s}$}{
        $X = X + \phi_{n_{s}}(y_{s_{t-1}}) - \phi_{n_{s}}(x_{t-1})$ \\}
        $x_{t-1} \leftarrow x_{t-1} + X$ \\ 
    }
  }

  \Return{$x_{0}$}
  
  \label{algorithm_label}
\end{algorithm}

In order to approximate this condition in every Markov transition during sampling, Choi et al. continuously refine the downsampled latent variable $x_{t}$ to be similar to the corresponding downsampled noisy version of reference image $y_{t}$, to ensure that both $x_{t}$ and $y_{t}$ share low frequency contents. $y_{t}$ is computed from reference image $y$ using Eq. 2 during sampling for every Markov transition. Specifically:

\vspace{-10pt}

\begin{align}
   p_{\theta}(x_{t-1}|x_{t}, c) \approx p_{\theta}(x_{t-1}|x_{t}, \phi_{N}(x_{t-1}) = \phi_{N}(y_{t-1})) 
\end{align}


\noindent where $\phi_{N}(\dots)$ is a low-pass linear filter (with N as the downsampling factor), and the term is approximated by ensuring the latent $x_{t-1}$ captures the missing low-frequency contents of $y_{t-1}$ after sampling from the unconditional DDPM.

\vspace{-10pt}

\begin{align}
    x_{t-1} = x_{t-1} + \phi_{N}(y_{{t-1}}) - \phi_{N}(x_{t-1})
\end{align}


We contend that by controlling the extent of low-pass filtering (factor N) of a linear filter $\phi$ for a given set of conditional or guidance images {$y_{1}, y_{2}, \dots, y_{m}$}, we can fine-tune our algorithm using a set of integers {$n_{1}, n_{2}, \dots, n_{m}$}. Here, each integer represents the extent of downsampling for a linear filter corresponding to each conditional image. This allows for valid image generation through a trained DDPM that shares low-level features (or similarity) with each of the conditional images. We modify Eq. 9 as:

\vspace{-10pt}

\begin{align}
    x_{t-1} = x_{t-1} + \sum_{s=1}^{M}(\phi_{n_{s}}(y_{s_{t-1}}) - \phi_{n_{s}}(x_{t-1}))
\end{align}

The downsampling factor $n_{s}$ for a conditional image will depend on the purpose and the nature of the conditional image in the generation of final images. In practice, the above strategy may only work well for a maximum of three or four conditional images. Fig. 1 visualizes our multi-conditional guidance and the steps in sampling where with each step the generated image gets closer to the desired super-subset of the image distribution. As is evident from the visualization; if integers \{$n_{1}, n_{2} \dots n_{m}$\} are not chosen carefully, there may not be a significant overlap between the subset distributions of conditional images in which case the image samplings may start generating inaccuracies in generated images. Steps 11 - 13 in Algorithm 1. illustrate the above process in our sampling of the synthetic lung CT images.



\section{Experiment Setup and Results}

We trained our DDPM \cite{nichol2021improved} on a dataset of (low-dose, 2D) lung CT-Scans of 5,000 patients. The images from the scans were extracted from the mid-abdomen regions, clearly showing the lungs along with the heart. The images were extracted in the entire relevant width of 2000 HU (-1000 HU to 1000 HU) for training our model; which enables the generation of images in the same HU range during the sampling process post training. That way, the images can be viewed at any HU window during their evaluation in our Visual Turing Test.


Figs. 2 and 3 showcase images generated with our model. Fig. 2 illustrates sets of guidance images, each comprising an anatomy map and a CT image. These sets serve as conditional (guidance) images for each of the 6 generated images, shown across the center bottom of the figure (see the caption for more detail). The results reveal that diverse anatomically accurate versions of a single CT image can be generated when annotations for the anatomy are available. Here, we generate anatomy maps using B-splines, as detailed in our prior work \cite{krishna2023}. It is noteworthy that this approach can be easily extended to simulate pathology/pathology types given a few annotated examples of CT images depicting such pathology. Fig. 3 presents the original full HU window generated images outlined with a red box, along with their decomposition in other clinically relevant windows. Visual inspections affirm their anatomical accuracy in each window, underscoring the effectiveness of DDPMs when assisted by guidance images in learning the nuances of anatomical structures across the entire HU range of lung CT images.

\begin{figure} [b!]
\vspace{-10pt}
\centering
\includegraphics[height=7.1cm]{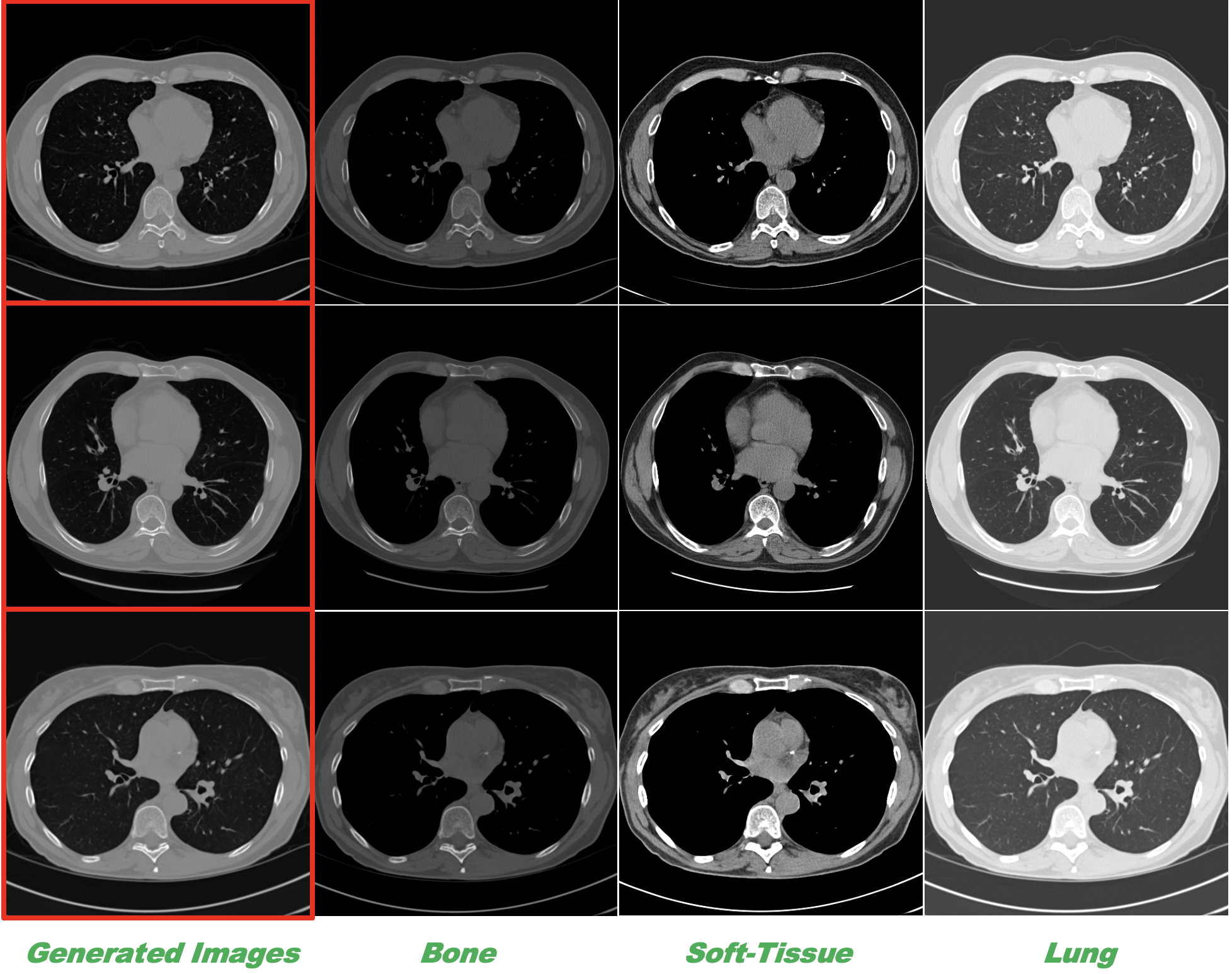}
\caption{Left-most column (outlined with a red box): images generated with our multi-conditional sampling algorithm, shown at full HU range. Other three columns: these images in their respective bone, soft-tissue, and lung windows.}
\label{fig}
\end{figure}

\subsection{Comparisons with Other Generative Models}

Table 1 shows comparative quantitative evaluations of a set of 10k generated full HU window lung CT images with the state-of-the-art image generative models namely NVIDIA's StyleGAN \cite{Karras2018ASG}, StyleGAN2 \cite{Karras2020} and PGGAN \cite{denton2015deep} that were trained on the same large dataset. We also compared these with the images generated from unguided sampling via the same trained DDPM to evaluate whether guided sampling has an effect on anatomical consistency apart from it being an annotated dataset generator. We chose the FID score because it measures the ”realism” of a set of generated images. 

The FID scores in Table 1 show that our model is at least almost as good quantitatively as the state-of-the-art StyleGAN2 if not better, and considerably better than the StyleGAN and the PGGAN. Additionally, unlike these models, our method is focused not only on just generating raw data but also its annotations. As such it could easily be expanded to generate CT scans with annotated pathology which is not possible in either of the above state-of-the-art models. 

We also performed an exhaustive set-level comparison where we gauged the Structural Similarity Index (SSIM) of the generated images against the large training dataset to measure the overall similarity of the generated images with respect to the training set images. As shown in Table 1, our model scores higher in SSIM than both of the StyleGANs. On visual inspection, the set of images generated via unconditional state-of-the-art-models can produce accurate generations but are prone to generating odd anatomies due to the absence of an anatomy controlled generation framework. This could explain at least partially the reason for the lower SSIM scores. Finally, our sampling strategy also outperforms the unguided sampled generations via the trained DDPM. 

\begin{table}[]
\caption{Comparing generative models.  }
\centering
{%
\begin{tabular}{|l|l|l|}
\hline
\textbf{} &
  \textbf{FID} &
  \textbf{Set-level SSIM}   \\ 
  \hline
\textbf{DiT}      & 82.83   & 0.38  \\ \hline
\textbf{StyleGAN}      & 81.57   & 0.31  \\ \hline
\textbf{StyleGAN2}      & 72.31    & 0.30   \\ \hline
\textbf{Unguided Sampling}      & 83.24   & 0.27   \\ \hline
\textbf{Guided Sampling}      & \textbf{69.85}   & \textbf{0.45}   \\ \hline

\end{tabular}%
}

\vspace{-10pt}
\end{table}

\subsection{Visual Turing Test}


\begin{figure}[b!]
\vspace{-15pt}
\centerline{\includegraphics[scale=0.24]{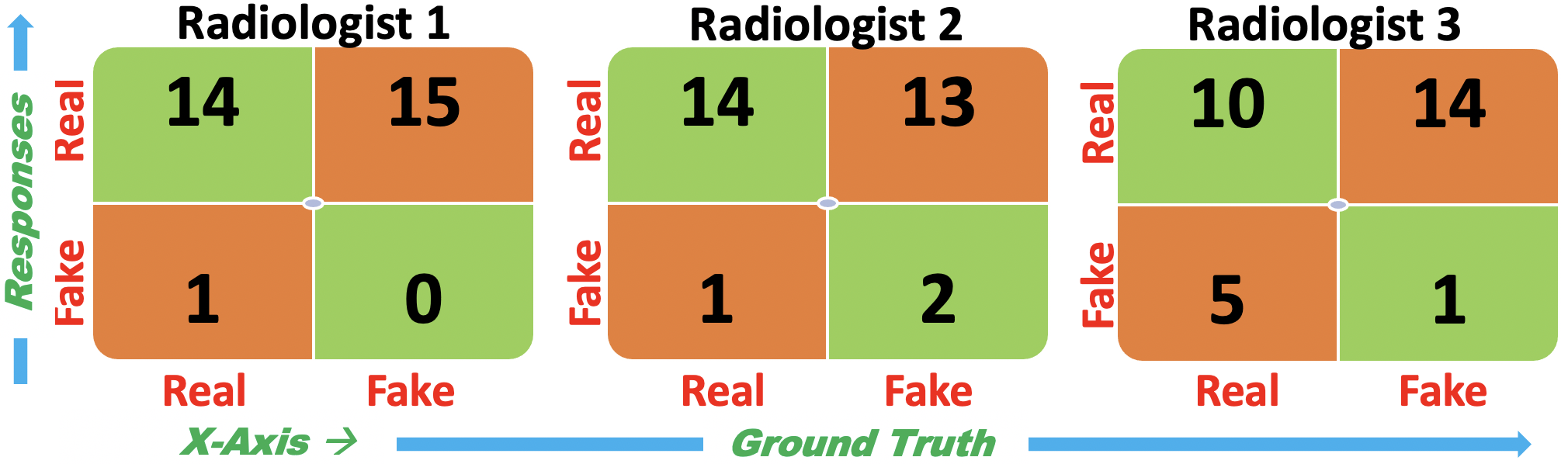}}
\vspace{-5pt}
\caption{Confusion matrices for the responses of the 3 radiologists. The overall accuracy of the responses is 45.56\% which is close to 50\%; a requirement for passing our Visual Turing Test. Proportion of 'True Negatives' (fake images identified as fake) is 6.67\% whereas proportion of 'False Negatives' (real images identified as fake) is 15.56\% }
\label{fig}
\end{figure}

We reran our Visual Turing Test \cite{krishna2023} with the assistance of three radiologists to evaluate the realism of our generated images. As previously, the test was administered to the radiologists by presenting them (via a web browser app) with a randomly selected lung CT image from a balanced set of 30 real and generated images, one at a time, in random order. The images were randomly chosen from bone, lung, and soft-tissue windows. Each image had two options: "Real" or "Fake." 

The test assesses if our model is able to generate medically accurate images. This is determined by measuring the number of times the model is able to fool the experts into thinking that a model generated image is a medical image obtained from a real patient. When experts are unable to separate the images into real or fake at least $50\%$ (chance baseline) of the time, the model is said to have passed the visual Turing test. The test was taken by the same three radiologists as in our previous study. Their responses are compiled in Fig. 4.  


The numbers in Fig. 4 indicate that the generative framework presented in this paper (unlike the one presented in our previous work \cite{krishna2023, krishna2023MP}) has passed the Visual Turing Test, as expert radiologists could not identify most of the fake (synthesized) lung CT images from the real ones. Upon analyzing the responses, we found that most responses were marked as 'Real' since the radiologists did not know that half of the shown images were 'Fake' and only marked an image as 'Fake' if they thought there was an anatomical/texture anomaly in the generated image. Even then, most of their responses labeled 'Fake' were, in fact, for the real images, showing that our guidance-based DDPM sampling scheme clearly passed the Visual Turing Test (see the caption for more details).

\section{Conclusions and Future Work}
Having demonstrated that our methodology can synthesize realist CT images with anatomical guidance, future work will extend this guidance to the synthesis of realistic pathology.


%

\section*{Acknowledgment}
NIH grant R01EB032716 funded this research. We
thank Dr. David Yankelevitz and Dr. Yeqing Zhu from the Icahn
School of Medicine at Mount Sinai, NY for image
evaluation.

\ifCLASSOPTIONcaptionsoff
  \newpage
\fi

\end{document}